\documentclass[conference]{IEEEtran}
\IEEEoverridecommandlockouts
\usepackage{cite}
\usepackage{stfloats}
\usepackage{caption}
\usepackage{subcaption}
\usepackage{amsmath,amssymb,amsfonts}
\usepackage{algorithmic}
\usepackage{graphicx}
\usepackage{textcomp}
\usepackage{tabularx}
\usepackage{multirow}
\usepackage{multicol}
\usepackage{booktabs}
\usepackage{listings}
\usepackage{float}
\usepackage{mathtools}
\usepackage{xcolor}
\usepackage{float}
\usepackage{flushend}

\usepackage{enumitem}
\usepackage [english]{babel}
\usepackage [autostyle, english = american]{csquotes}
\MakeOuterQuote{"}

\def\BibTeX{{\rm B\kern-.05em{\sc i\kern-.025em b}\kern-.08em
    T\kern-.1667em\lower.7ex\hbox{E}\kern-.125emX}}

\usepackage{siunitx}
\usepackage{caption}
\usepackage{subcaption}
\begin{document}
\setlength{\parskip}{0pt}

\UseRawInputEncoding
\title{\LARGE \bf
Force-Aware Autonomous Robotic Surgery*}

\author{Alaa Eldin Abdelaal$^{1}$ \textit{Member, IEEE}, Jiaying~Fang$^{2}$, Tim~N.~Reinhart$^3$, Jacob~A.~Mejia$^3$, Tony~Z.~Zhao$^3$, \\ Jeannette~Bohg$^3$ and Allison~M.~Okamura$^1$ \textit{Fellow, IEEE}
\thanks{This work has been funded in part by the Stanford Institute for Human-Centered Artificial Intelligence, Intuitive Surgical, Inc., and a postdoctoral fellowship from the Natural Sciences and Engineering Research Council of
Canada (NSERC) to Alaa Eldin Abdelaal. Any research, findings, conclusions, or recommendations expressed in this work are those of the authors, and not of Intuitive Surgical.}
\thanks{$^{1}$A.~E.~Abdelaal and A.~M.~Okamura are with the Mechanical Engineering Department, Stanford University, 440 Escondido Mall Building 530, Stanford, CA 94305.
        {\tt\small (email: abdelaal@stanford.edu)}}%
\thanks{$^{2}$J.~Fang is with the Electrical Engineering Department, Stanford University, 350 Jane Stanford Way, Stanford, CA 94305.}        
\thanks{$^{3}$T.~N.~Reinhart, J.~A.~Mejia, T.~A.~Zhao and J.~Bohg are with the Department of Computer Science, Stanford University, 353 Jane Stanford Way, Stanford, CA 94305.}
}

\maketitle
\thispagestyle{empty}
\pagestyle{empty}

\begin{abstract}
This work demonstrates the benefits of using tool-tissue interaction forces in the design of autonomous systems in robot-assisted surgery (RAS). Autonomous systems in surgery must manipulate tissues of different stiffness levels and hence should apply different levels of forces accordingly. We hypothesize that this ability is enabled by using force measurements as input to policies learned from human demonstrations. To test this hypothesis, we use Action-Chunking Transformers (ACT) to train two policies through imitation learning for automated tissue retraction with the da~Vinci Research Kit (dVRK). To quantify the effects of using tool-tissue interaction force data, we trained a "no force policy" that uses the vision and robot kinematic data, and compared it to a "force policy" that uses force, vision and robot kinematic data. When tested on a previously seen tissue sample, the force policy is 3 times more successful in autonomously performing the task compared with the no force policy. In addition, the force policy is more gentle with the tissue compared with the no force policy, exerting on average 62\% less force on the tissue. When tested on a previously unseen tissue sample, the force policy is 3.5 times more successful in autonomously performing the task, exerting an order of magnitude less forces on the tissue, compared with the no force policy. These results open the door to design force-aware autonomous systems that can meet the surgical guidelines for tissue handling, especially using the newly released RAS systems with force feedback capabilities such as the da~Vinci~5. 

\end{abstract}


\section{INTRODUCTION}
        \label{sec:introduction}
        Robot-assisted surgery (RAS) has had considerable success with well over 10 million procedures conducted using the da~Vinci Surgical System alone, the most widely used surgical robotics platform worldwide. It has been used in more than 2.2 million procedures in 2023 and it has an installed base of more than 9500 systems in many countries around the world~\cite{intuitive}. RAS has been used in several surgical specialties such as urology~\cite{autorino2020robotic} and gynecology~\cite{kristensen2017robot}.

Many surgical robotic platforms include three robotic manipulators; two are usually used for the main part of the task, and a third one is often used to perform several auxiliary tasks~\cite{sundaram2005utility, rogers2009maximizing}. Because a surgeon can only control two arms at a time in RAS, they must switch the robot's control system to use the third arm to perform one part of the task and then switch the control system back to manipulate the other two arms as needed. This switching increases the cognitive load of the surgeon~\cite{liu2015review} and can cause negative outcomes such as collisions between the three robot arms, unnecessarily prolonging the procedure, and tearing of the human tissue~\cite{catchpole2016safety}. This provides a huge opportunity to develop systems that can autonomously carry out the tasks of the third robotic arm.

An important category of tasks carried out using the third robotic arm requires the physical interaction with tissue. This includes tasks such as tissue retraction, where the goal is to grasp a tissue flap and move it to uncover an area of interest underneath it~\cite{steele2013current}.

The tool-tissue interaction forces are important for the successful execution of these tasks. In surgical practice, applying too little force can hinder the task execution, and applying too much force can damage the tissue, with potentially serious health consequences~\cite{putzer2015retracting}. In addition, in any surgical procedure, surgeons operate on tissue samples of different stiffness levels. The forces exerted onto these tissue samples differ based on their stiffness~\cite{golahmadi2021tool}, and surgical guidelines recommend applying appropriate forces to each tissue type based on its characteristics~\cite{poulose1999human}. 

The goal of this work is to design and evaluate \textit{force-aware} systems to automate the motion of the third robotic arm in RAS. By \textit{force-aware}, we mean that tool-tissue interaction forces are part of the autonomous system design so that it can manipulate tissues of different stiffness levels. The contributions of this work are as follows:
\begin{itemize}[noitemsep,topsep=0pt]
\item We develop a force-aware autonomous system for RAS, using robot learning from demonstrations, to automate the motion of the third robotic arm of the da Vinci Surgical System.
\item We show that our force-aware autonomous system is three times more successful in autonomously performing a tissue retraction task compared with a force-agnostic system.
\item We show that our force aware autonomous system is more gentle with the tissue compared with the force-agnostic one, by comparing the applied forces by the two systems as they autonomously carry out the task.
\item We show that our force-aware autonomous system can generalize to an unseen tissue sample of different stiffness level. In particular, the force-aware system is three and half (3.5) times more successful on the unseen tissue sample than the force-agnostic one. In addition, the force-aware system is more gentle with the unseen tissue compared with the force-agnostic one. 
\end{itemize}
        
\section{RELATED WORK}
        \label{sec:related_works}
        Many groups have developed methods and algorithms for autonomous robotic surgery for surgical tasks~\cite{attanasio2021autonomy}. The main goal of this work is to automate repetitive tasks in surgery so that the surgeon can focus more on the demanding parts of the surgical procedure~\cite{yip2019robot}. For example, Shademan et al.~\cite{shademan2016supervised} built an autonomous system to perform various types of the suturing task on soft tissue that was tested successfully ex vivo and in vivo. Furthermore, Padoy and Hager~\cite{padoy2011human} developed a human-machine collaborative system (HMCS) to perform suturing where parts of the task are performed by a human and the remaining parts of the task are performed autonomously by a surgical robot arm. Several other autonomous systems have been proposed for suturing, such as~\cite{saeidi2022autonomous, pedram2020autonomous, abdelaal2021parallelism, ostrander2024current}. Autonomous algorithms have also been developed for other tasks such as blood suction~\cite{richter2021autonomous}, intracardiac catheter navigation~\cite{fagogenis2019autonomous}, brain tumor ablation~\cite{hu2015semi}, tissue retraction \cite{patil2010toward, nagy2018surgical, jansen2009surgical, attanasio2020autonomous}, and moving the surgical camera/endoscope~\cite{ma2019autonomous}, to name a few. Some of these algorithms are approved by the U.S. Food and Drug Administration (FDA) and are commercially available, e.g.,~\cite{wijsman2018first}.

One of the most successful techniques to automate surgical tasks is Learning from Demonstration (LfD) or Imitation Learning~\cite{argall2009survey}. The main idea of LfD is to teach robots how to perform a task based on data collected from human demonstrations. For example, Van den Berg et al.~\cite{van2010superhuman} developed a system to teach a surgical robot to autonomously perform a two-handed knot tying task based on expert demonstrations. Other examples include~\cite{murali2015learning, su2021toward, schwaner2021autonomous, pore2021learning, osa2014trajectory, shin2019autonomous, rivas2019transferring}. More recently, Kim et al.~\cite{kimsurgical} successfully tested two state-of-art imitation learning algorithms (namely Action Chunking Transformers~\cite{zhao2023learning} and Diffusion Policy~\cite{chi2023diffusionpolicy}) to automate several surgical tasks on the da~Vinci Surgical System.

This body of work does not consider the forces exerted by the surgical tools on the tissue in the design of the proposed LfD pipelines to automate surgical tasks. Instead, previous work has used a combination of motion/kinematic data from the surgical tools and video data of the surgical scene as inputs. This contrasts with the vast literature in RAS on estimating or sensing tool-tissue interaction forces~\cite{black20206, 10183676, chua2020toward} and using them to provide haptic feedback for surgeons (to improve their performance) in fully teleoperated surgical systems~\cite{abdi2020haptics, koehn2015surgeons, 8469020}. This is also in opposition to surgical guidelines, which recommend applying appropriate forces to each tissue type based on its characteristics~\cite{poulose1999human}.

The goal of this study is to fill this gap. In particular, we demonstrate the benefits of considering forces in designing autonomous systems in RAS. We measured the effect of using forces on the following: (i) The successful autonomous execution of tasks in RAS, similar to other applications in the literature~\cite{lee2020making, zhang2023visual}; (ii) How gentle the autonomous system is on the tissue, which is an important consideration in RAS; and (iii) The ability of the autonomous system to generalize its behavior to unseen tissue samples, another important practical consideration in RAS.
        
\section{METHODS}
        \label{sec:approach}
Our approach uses imitation learning to automate the motion of a robotic manipulator to perform a RAS task based on collected demonstration data from an expert performing the same task. We then evaluate the performance of the resulting autonomous system with and without the collected force data during demonstrations. We evaluate the system on the same tissue sample used for data collection and also on an unseen tissue sample to test the generalization of the developed system. In the following subsections, we describe the different components of the proposed method.  

\subsection{Imitation Learning Network}
\label{sec:network}

We adapted the action-chunking-transformer (ACT) architecture proposed in~\cite{zhao2023learning} to fit this work. This architecture has been shown to be successful in several fine-grained manipulation tasks, based on demonstration data collected from teleoperation robotic platforms. This includes surgical tasks using the da~Vinci Surgical System as in~\cite{kimsurgical}.

\begin{figure}[!b]
    \centering
    \includegraphics[width=\linewidth]{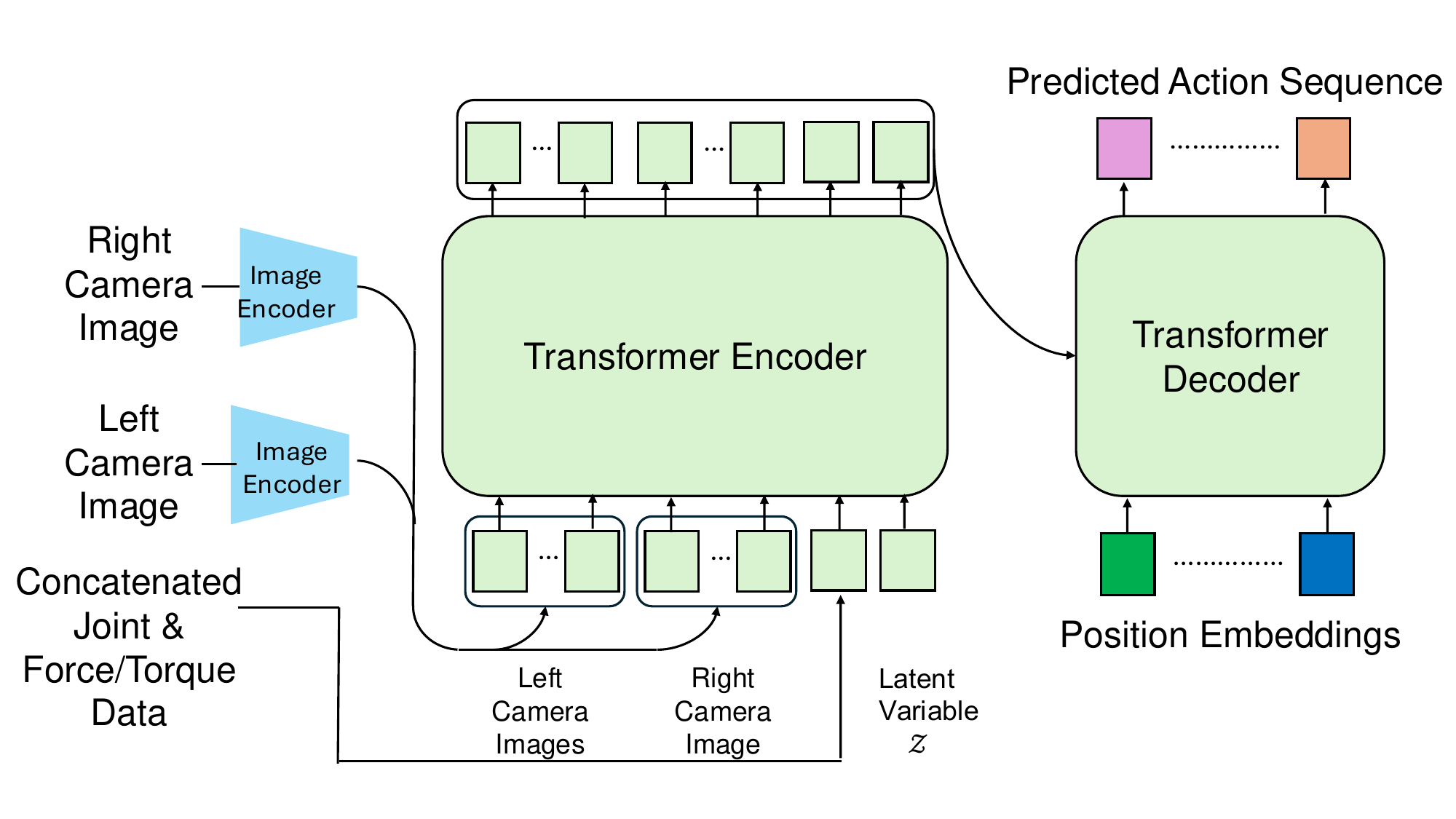}
    \caption{The CVAE decoder used to generate the learned policy in the modified ACT architecture. The CVAE decoder synthesizes the stereoscopic images, concatenated robot joints and force/torque data, and latent variable $z$ with a transformer encoder. Then, it predicts the robot's action sequence with a transformer decoder.}
    \label{fig:modified_ACT_arch}
\end{figure}

\begin{figure*}[!b]
    \centering
    \centering
    \begin{subfigure}[t]{0.42\textwidth}
         \centering
         \includegraphics[width=\textwidth]{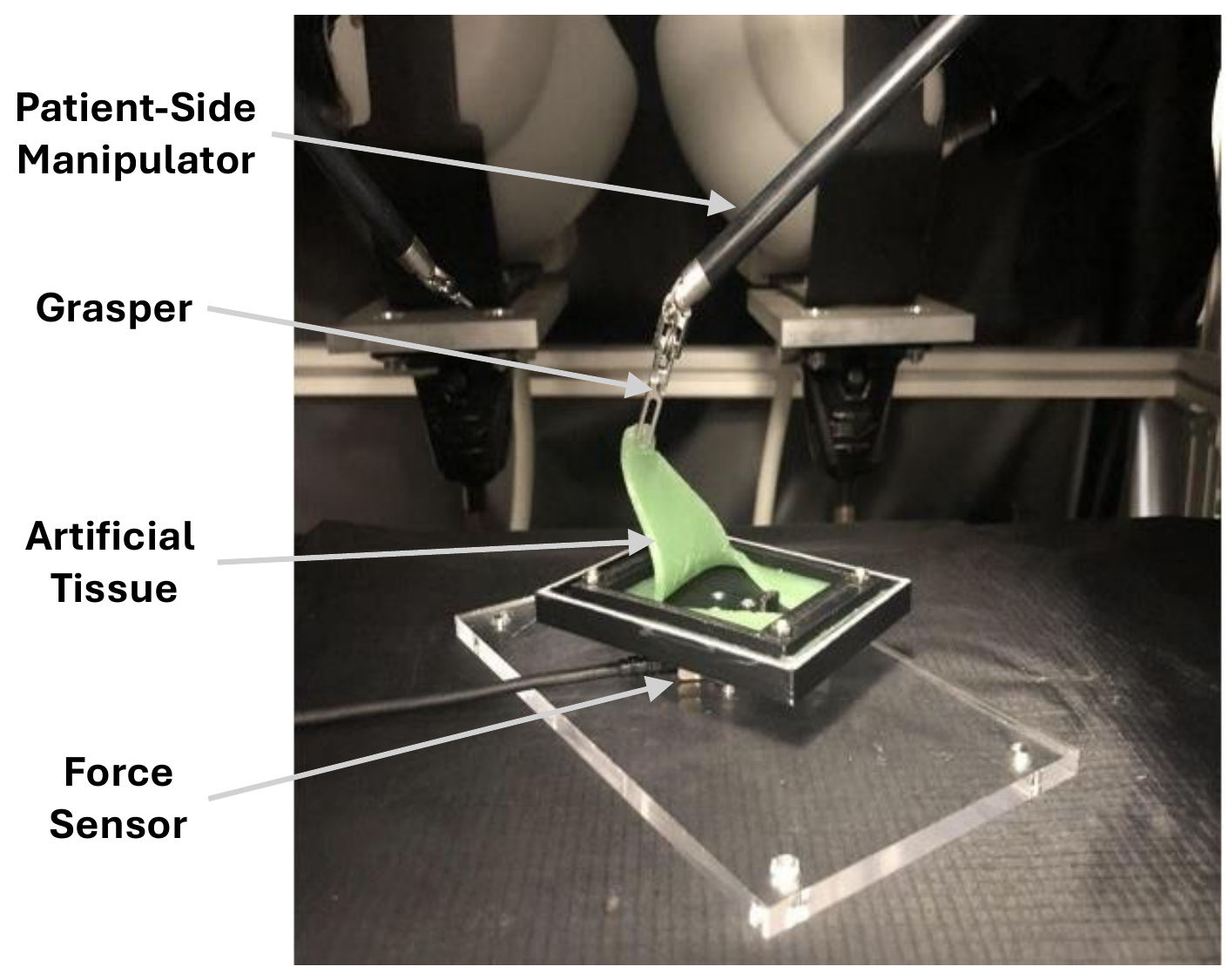}
         \caption{(a) }
         \label{fig:setup}
     \end{subfigure}
     \hspace{30pt}
     \begin{subfigure}[t]{0.3\textwidth}
         \centering
         \includegraphics[width=\textwidth]{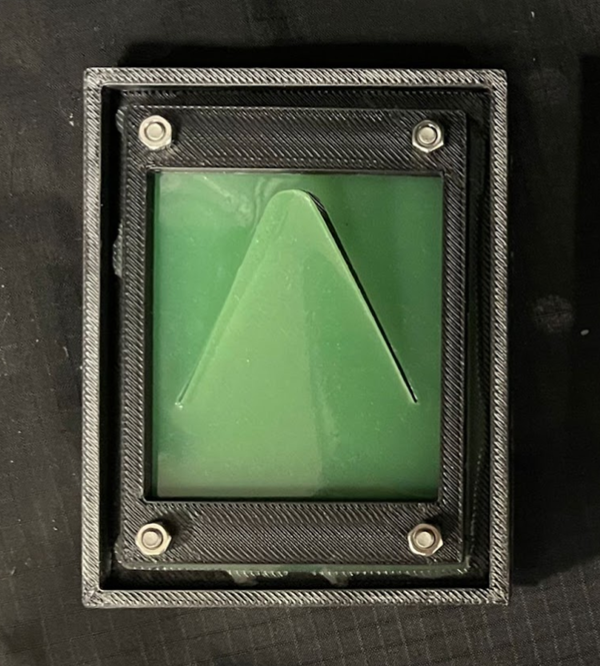}
         \caption{(b) }
         \label{fig:tissue_sample}
     \end{subfigure}
     \caption{(a) The hardware setup used in this work. The tissue sample is mounted on a rigid board. The force/torque sensor is mounted under the board and is placed on a flat surface. This figure shows the tissue being lifted as the final part of the tissue retraction task. (b) One of the tissue samples used in this work. The task is to retract the triangular area to uncover the area underneath it.}
     \label{fig:whole_setup}
\end{figure*}

The architecture of the ACT consists of a conditional variational autoencoder (CVAE). The policy is trained as a generative model that generates actions conditioned on current observations. The CVAE includes a CVAE encoder and decoder. The CVAE encoder is only used during training time. Its architecture is implemented as a transformer encoder. For computational efficiency, the CVAE encoder only takes low-level observations (e.g., robot's joint kinematics) and predicts the mean and variance of the latent variable's ($z$) distribution. The CVAE decoder serves as the policy during evaluation. It consists of a transformer encoder and decoder. In the original ACT, the observations to the transformer encoder include vision data (i.e., images from the teleoperation system's camera(s)) and kinematic data from the robot (i.e., robot's joint angles). The images are first processed by ResNet18 image encoders~\cite{7780459} before inputting them to the transformer. The current observations and latent variable $z$ serve as the inputs to the transformer encoder of the CVAE decoder. At test time, the latent variable $z$ is set to be the mean of the prior (i.e., zero). The CVAE decoder predicts the next $k$ actions in the robot's joint space. The chunk size $k$ is chosen to match the control frequency of the used robotic platform. To make the learned policy more reactive, the model is queried at every time step, then takes the weighted average of the predicted actions as the final action for the robot.

The adapted ACT architecture for our work is shown in Fig.~\ref{fig:modified_ACT_arch}. We include the tool-tissue interaction forces/torques as part of the input. This includes additional six input values representing the forces and torques exerted by the surgical instrument on the tissue in six degrees of freedom (DoF). The original architecture does not use force data. The modified architecture gets vision input from the surgical robotic platform's stereoscopic camera only, instead of four cameras as in the original architecture. This is in line with the visual feedback provided in real robot-assisted surgeries. Moreover, the action data in the modified architecture come from the follower robotic manipulators, instead of the leader robotic manipulators in the original architecture. This is because the kinematic configuration of the leader and follower manipulators are not necessarily the same in surgical robotic platforms as in the hardware platform where the original architecture was tested.

\subsection{Hardware Setup}
\label{sec:setup}
The hardware setup of the data collection and experiments in this work is shown in Fig.~\ref{fig:setup}. We used the da~Vinci Research Kit (dVRK)~\cite{6907809}, on the first generation da~Vinci System, as our surgical robotic platform in this work. A python script is used to log the joint angles of the dVRK manipulators along with the dVRK's stereoscopic camera. An ATI 6-axis force/torque sensor is mounted underneath the task setup and is used to collect the force data. In practice, the force data can be collected by commercial available surgical robotics platforms with force feedback capabilities such as the da~Vinci~5 (Intuitive Surgical Inc.) and Hugo (Medtronic Inc.).

\subsection{Task}
\label{sec:task_design}

We applied the imitation learning model to a tissue retraction task, where the goal is to grasp a tissue flap and move it to uncover an area of interest underneath it~\cite{steele2013current}. It is a task that is carried out using the surgical robot's third arm and it requires direct physical interaction between the tissue and surgical instrument.  

The task is considered completed successfully if all of the following three conditions are satisfied: 
\begin{enumerate}
    \item[(a)] The tissue is grasped correctly and the area beneath it is uncovered.
    \item[(b)] The tissue remains lifted for a reasonable amount of time and the robotic manipulator does not drop the tissue at any point until the end of the task.
    \item[(c)] The tissue sample is not damaged due to excessive force exerted on it.
\end{enumerate}

We created two tissue samples of different stiffness levels, that look visually the same, to be used in this work. The less stiff tissue sample has a 100\% modulus of 55 kPa and the more stiff one has a 100\% modulus of 151 kPa. The less stiff tissue sample is used for data collection and the more stiff sample is only used to test whether the proposed approach can generalize to unseen tissue samples. The two tissue samples were created using two different silicone materials (i.e., Dragon Skin and Ecoflex) from Smooth-On, Inc. The two samples were created in the same rectangular mold. The mold was designed so that the resulting samples have a triangular area as shown in Fig.~\ref{fig:tissue_sample}, which is grasped and lifted to uncover the area underneath it during task execution.

\subsection{Data Collection and Training}
\label{sec:data_collection}
\begin{figure*}[!t]
    \centering
    \begin{subfigure}[t]{0.4\textwidth}
         \centering
         \includegraphics[width=\textwidth]{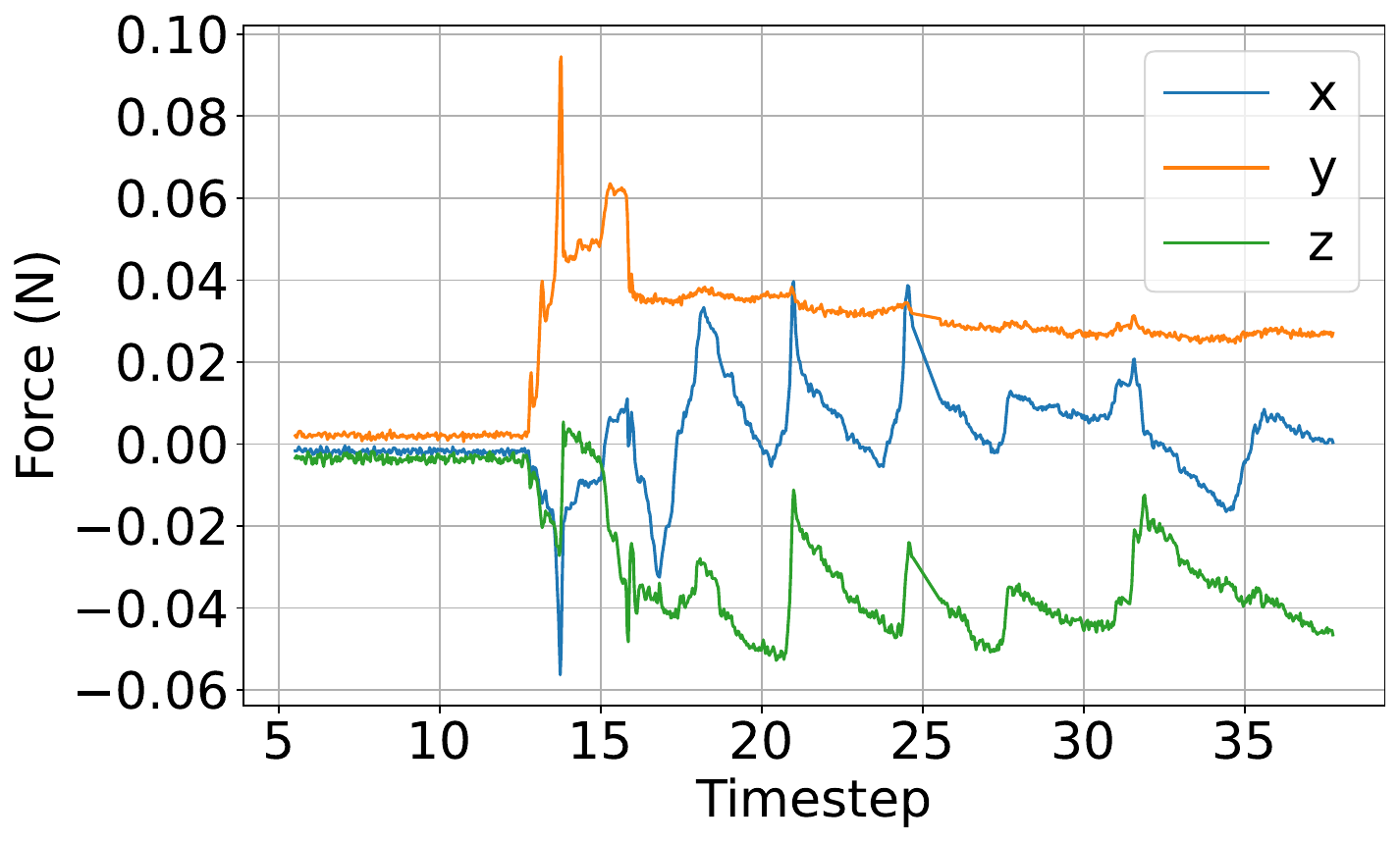}
         \caption{(a)}
         \label{fig:smooth_force}
     \end{subfigure}
     \begin{subfigure}[t]{0.55\textwidth}
         \centering
         \includegraphics[width=\textwidth]{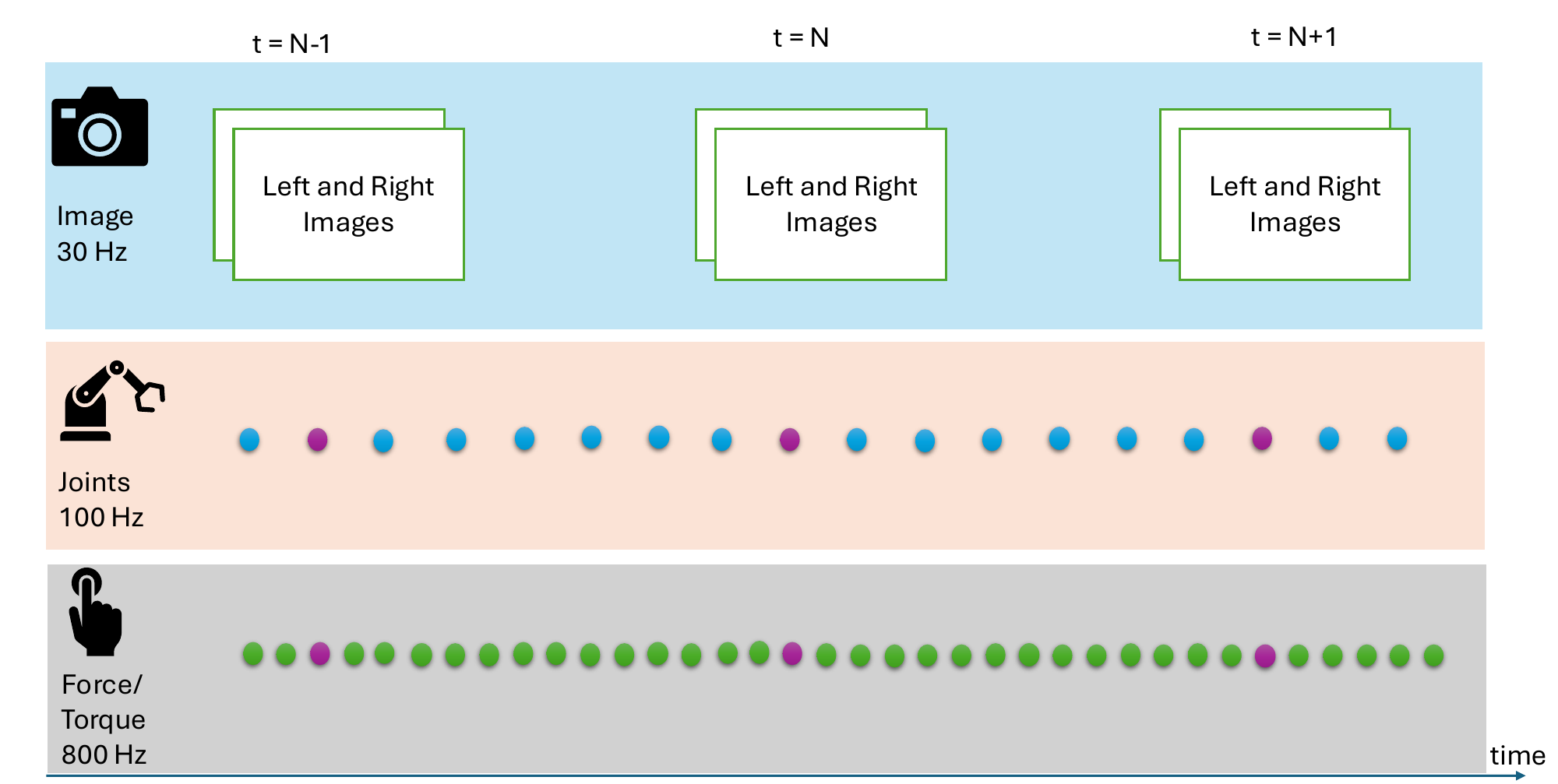}
         \caption{(b)}
         \label{fig:sync_vis}
     \end{subfigure}
    \caption{\textbf{Data Collection Scheme}: (a) Visualization of smoothed force values. The raw force/torque data was smoothed by applying an averaging window to reduce the effect of the measurement noise. (b) Visualization of data collection of multiple modalities. All modalities were recorded at the camera frequency: 30 Hz. The purple dots indicate the force/torque values (smoothed) and joint kinematics recorded at time steps N-1, N, and N+1.}
    \label{fig:data_collection}
\end{figure*}
We collected 60 demonstrations of the tissue retraction task from a human expert teleoprating the dVRK. The position, orientation and height of the tissue sample were randomized across the collected demonstrations. The human expert was instructed to complete tissue retraction as gently as possible, imitating gentle tissue manipulation in real surgeries. The same tissue sample was used throughout the entire data collection process.

We collected three types of data in each demonstration. The joint angles of the dVRK's patient side manipulators (PSMs) were collected at 100 Hz. The 6 DoF force/torque data from the ATI force sensor were collected at 800 Hz. The left and right images from the dVRK's camera were collected at 30 Hz. In the final data set, we downsampled the joint and force/torque data to match the recording frequency of the images. The raw force/torque data was smoothed first by applying an averaging window to reduce the effect of the measurement noise, as shown in Fig.~\ref{fig:smooth_force}. We then downsampled the resulting data to 30 Hz. 

The general data collection scheme is shown in Fig.~\ref{fig:sync_vis}. We represent the observation (i.e., the input to the imitation learning architecture) during one time step as follows:
\begin{equation}
    o_I = [I_{\text{left}}, I_{\text{right}}],
\end{equation}
\begin{equation}
    o_f = [f_x, f_y, f_z, \tau_x, \tau_y, \tau_z],
\end{equation}
\begin{equation}
    o_q = [q^{\text{PSM1}_i}, q^{\text{PSM2}_i}], \quad i = 1, \ldots, 7,
\end{equation}
where $o_I$ is the image observations, which includes both left and right images. $o_f$ is the downsampled force/torque observation, which is in $\mathbb{R}^6$. $o_q$ is the joint angles of the left patient side manipulator (PSM1) and the right one (PSM2), which are in $\mathbb{R}^{14}$. We represent the action (i.e., the output of the imitation learning architecture) during one time step as follows:
\begin{equation}
    a = [qdes^{\text{PSM1}}_i, qdes^{\text{PSM2}}_i], \quad i = 1, \ldots, 7,
\end{equation}
where $qdes^{\text{PSM1}}_i$ and $qdes^{\text{PSM2}}_i$ are the joint actions (i.e., the predicted joint values) at PSM1 and PSM2 joints. The action is also in $\mathbb{R}^{14}$. 

The collected demonstrations were used to train two policies to automate the motion of the robotic manipulator to perform the tissue retraction task. The first policy was generated using the network in Fig.~\ref{fig:modified_ACT_arch}. We refer to this policy as the \textit{force policy}. The second policy was generated using another version of the same network that only uses vision and joint kinematic data as inputs. This version of the network did not use the force data. We refer to this second policy as the \textit{no force policy}.   
The two policies were trained for 20,000 epochs. The training parameters we used are shown in Table \ref{tab:training_parameters}.

\begin{table}[!b]
\centering
\begin{tabular}{c|c}
\hline
\textbf{H-params} & \textbf{Values}  \\ \hline
\# of demonstrations & 60\\
epochs & 20,000\\
learning rate & 1e-5\\
batch size & 8\\
\# encoder layers & 4\\
\# decoder layers & 7\\
feedforward dimension & 3200\\
hidden dimension & 512\\
\# heads & 8\\
chunk size & 100
\end{tabular}
\caption{The hyper-parameters for training the ACT policy.}
\label{tab:training_parameters}
\end{table}

\subsection{Hypotheses, Evaluation and Performance Metrics}
\label{sec:eval_metrics}
We hypothesize that the force policy will have a higher success rate in the tissue retraction task compared with the no force policy. In addition, we hypothesize that the force policy will be more gentle with the used tissue samples compared to the no force policy. We also hypothesize that the force policy will better generalize to a previously unseen tissue sample than the no force policy. 

The proposed approach is tested to evaluate the effect of using the force/torque data on the following:
\begin{enumerate}
    \item[(a)] The success rate in performing the tissue retraction task autonomously.
    \item[(b)] The applied forces on the tissue during autonmous task execution.
    \item[(c)] The generalization to an unseen tissue sample.
\end{enumerate}

Towards this end, we tested the two learned policies by rolling them out on the dVRK to autonomously perform the tissue retraction task on the same tissue used for training. Each policy was rolled out 50 times. We randomized the position, orientation and height of the tissue sample in each roll out. We recorded the success/failure of each roll out. A successful roll out satisfies the three conditions in Section~\ref{sec:task_design} above. 

In addition, we collected the tool-tissue interaction forces in each roll out for further analysis using the same ATI force/torque sensor used for data collection. In order to compare the force values more intuitively, we calculated the $l_2$ norm of the force vector $\mathbf{f}$ as the total exerted force:
\begin{equation}
    f_{\text{total}} = \left\Vert \mathbf{f} \right\Vert_2,
\end{equation}
where $\mathbf{f}$ contains the force values recorded from the sensor:
\begin{equation}
\mathbf{f} = \begin{bmatrix}
    f_x\\
    f_y\\
    f_z
\end{bmatrix}.
\end{equation}
\\

\section{RESULTS}
        \label{sec:results}
        \subsection{Seen Tissue}
\label{sec:Results_on_Seen_tissue}

\begin{table*}[!b]
\centering
\begin{tabular}{ccccc}
\hline
\textbf{Policy} & \textbf{Mean Force (N)} & \textbf{Std. Dev. of Force(N)} & \textbf{Max Force (N)} & \textbf{Success Rate} \\ \hline
Force Policy (All Rollouts)& 0.29 & 0.54 & 6.65 & \multirow{2}{*}{\textbf{76\%}}\\ 
Force Policy (Successful Rollouts)& 0.26 & 0.43 & 6.65 & \\ 
\hline
No Force Policy (All Rollouts)& 0.47 & 0.8 & 7.08 & \multirow{2}{*}{26\%}\\
No Force Policy (Successful Rollouts)& 0.28 & 0.51 & 7.08 & 
\end{tabular}
\caption{The results of the policy roll outs on the seen tissue sample. }
\label{tab:seen_tissue_statistics}
\end{table*}

\begin{figure*}[!b]
    \centering
    \begin{subfigure}[b]{0.32\textwidth}
        \centering
        \includegraphics[width=\textwidth]{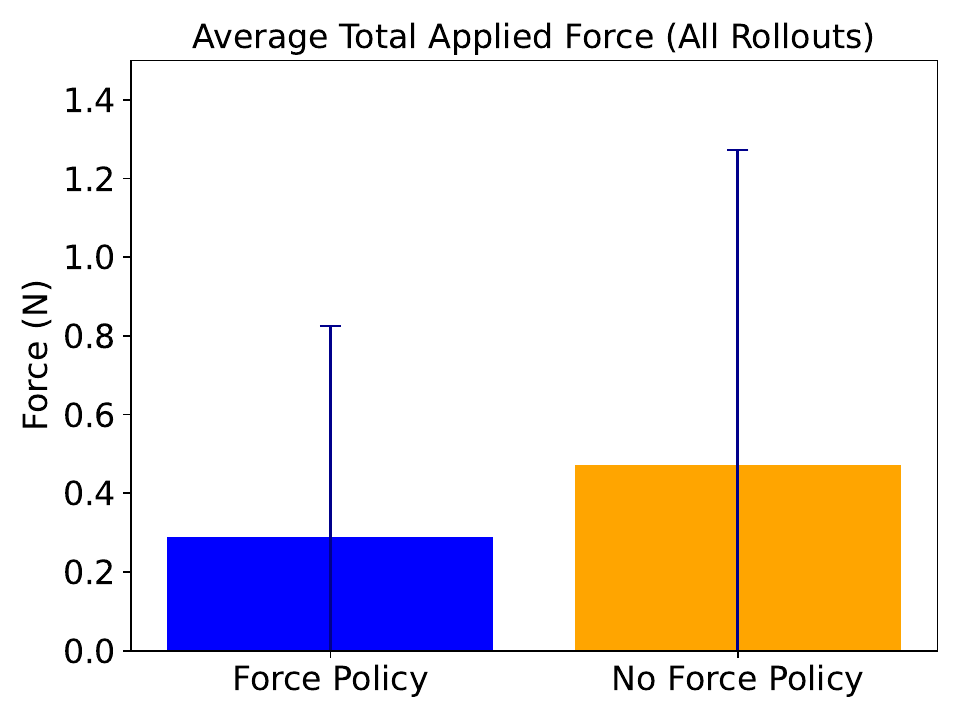}
        \caption{(a)}
        \label{seen_tissue_bar_plot}
    \end{subfigure}
    \hfill
    \begin{subfigure}[b]{0.32\textwidth}
        \centering
        \includegraphics[width=\textwidth]{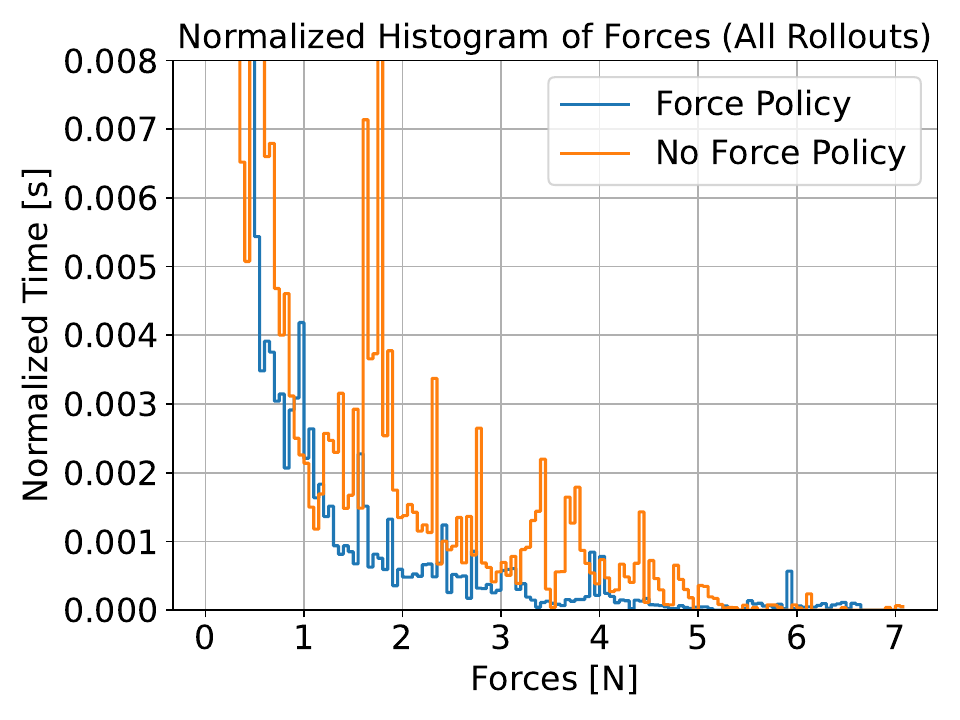}
      \caption{(b) }
      \label{seen_tissue_histogram_forces_all_rollouts}
    \end{subfigure}
    \hfill
    \centering
    \begin{subfigure}[b]{0.32\textwidth}
        \centering
        \includegraphics[width=\textwidth]{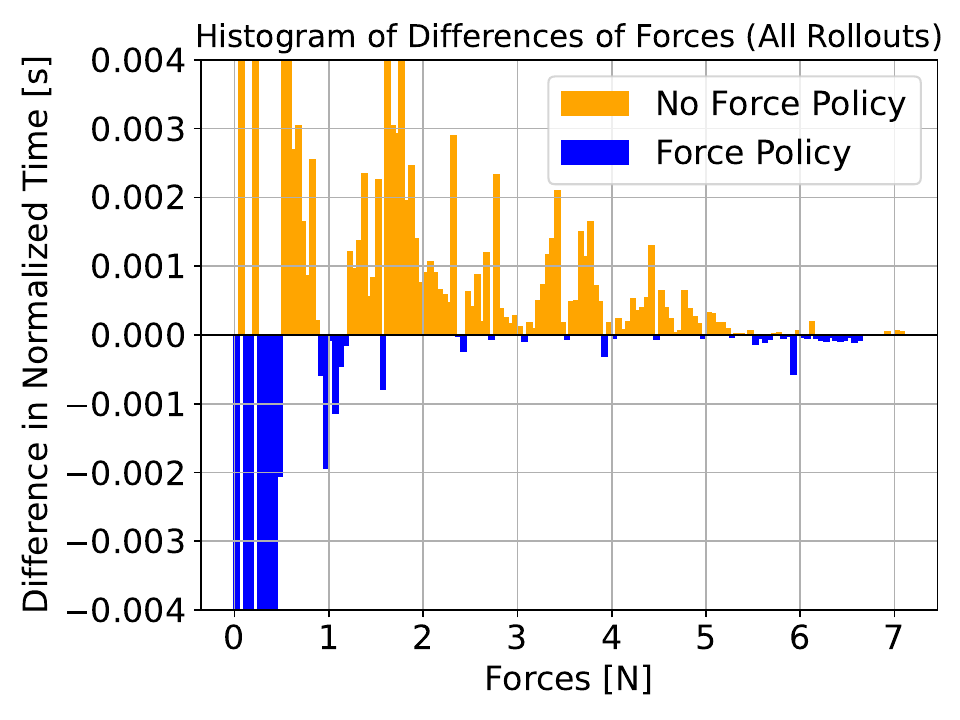}
      \caption{(c) } \label{seen_tissue_histogram_difference_forces_all_rollouts}
    \end{subfigure}
    \caption{(a) Average total applied force and standard deviation for both policy roll outs on the seen tissue sample. (b) Normalized histogram of forces showing the total time a specific force is applied during both policy roll outs on the seen tissue sample. (c) Histogram of differences of applied force during both policy roll outs on the seen tissue sample.}
\end{figure*}

Our results in Table~\ref{tab:seen_tissue_statistics} show that our force-aware autonomous system is three times more successful than the force-agnostic system on the tissue retraction task on the seen tissue sample. Over the 50 roll outs of each policy, the force policy was successful 76\% of the time (38 successful task executions out of 50 roll outs), and the no force policy was successful only 26\% of the time (13 successful task executions out of 50 roll outs).

Furthermore, a comparison between the force policy and no force policy during roll outs reveals distinct differences in the applied forces by the two policies. As displayed in Fig.~\ref{seen_tissue_bar_plot} and Table~\ref{tab:seen_tissue_statistics}, the force policy roll outs resulted in a lower mean force applied to the tissue, and also a lower standard deviation, compared with the no force policy. In particular, the no force policy applies 62\% more force on average than the force policy calculated across all roll outs. This suggests that the application of force is smoother and more consistent when force data is utilized. This result is statistically significant. The statistical analysis of the above result using a t-test revealed a t-statistic of $-30.5$ and a p-value $<$ 0.01.

The normalized histogram of forces in Fig.~\ref{seen_tissue_histogram_forces_all_rollouts} shows the distribution of the duration for which different forces were applied during the policy roll outs. The force policy demonstrates a steeper decline in the duration of force application as the force increases, indicating a more delicate and controlled interaction with the tissue. Conversely, the no force policy exhibits longer durations for higher forces, suggesting a more aggressive application of force. This behavior implies that the force policy is more effective in minimizing the application of large forces, which confirms that the force policy interacts in a smoother and safer way with the tissue. 

This is also illustrated in Fig.~\ref{seen_tissue_histogram_difference_forces_all_rollouts}, which shows the difference of the time a given force is applied by the two policies. If all bars in the plot were orange, the no force policy would apply every given force more often than the force policy. Because the bars are predominantly orange, especially for forces larger than 1N, we can conclude again that the no force policy applies large forces more often and that the force policy is more gentle in handling the tissue compared with the no force policy. This represents a desired behavior in surgery. 

\subsection{Unseen Tissue}
\label{sec:Results_on_the_Unseen_Tissue}
\begin{table*}[!b]
\centering
\begin{tabular}{cccccc}
\hline
\textbf{Policy} & \textbf{Mean Force (N)} & \textbf{Std. Dev. of Force (N)} & \textbf{Max Force (N)} & \textbf{Success Rate} \\ \hline
Force Policy (All Rollouts)& 0.4 & 0.91 & 7.62 & \multirow{2}{*}{\textbf{70\%}} \\ 
Force Policy (Successful Rollouts)& 0.39 & 0.86 & 7.62 & \\ \hline
No Force Policy (All Rollouts)& 0.84 & 1.37 & 8.68 & \multirow{2}{*}{20\%}\\
No Force Policy (Successful Rollouts)& 0.42 & 0.87 & 8.68&\\
\end{tabular}
\caption{The results of the policy roll outs on an unseen tissue sample. }
\label{tab:generalization_statistics}
\end{table*}

Our results in Table~\ref{tab:generalization_statistics} show that our force-aware autonomous system is three and half (3.5) times more successful than the force-agnostic system on the tissue retraction task on the unseen tissue sample. The success rate of each policy slightly dropped compared with their individual performance on the seen tissue, as expected. Nevertheless, over the 50 roll outs of each policy, the force policy was successful 70\% of the time (35 successful task executions out of 50 roll outs), and the no force policy was successful only 20\% of the time (10 successful task executions out of 50 roll outs).

\begin{figure*}[!b]
    \centering
    \begin{subfigure}[b]{0.32\textwidth}
        \centering
        \includegraphics[width=\textwidth]{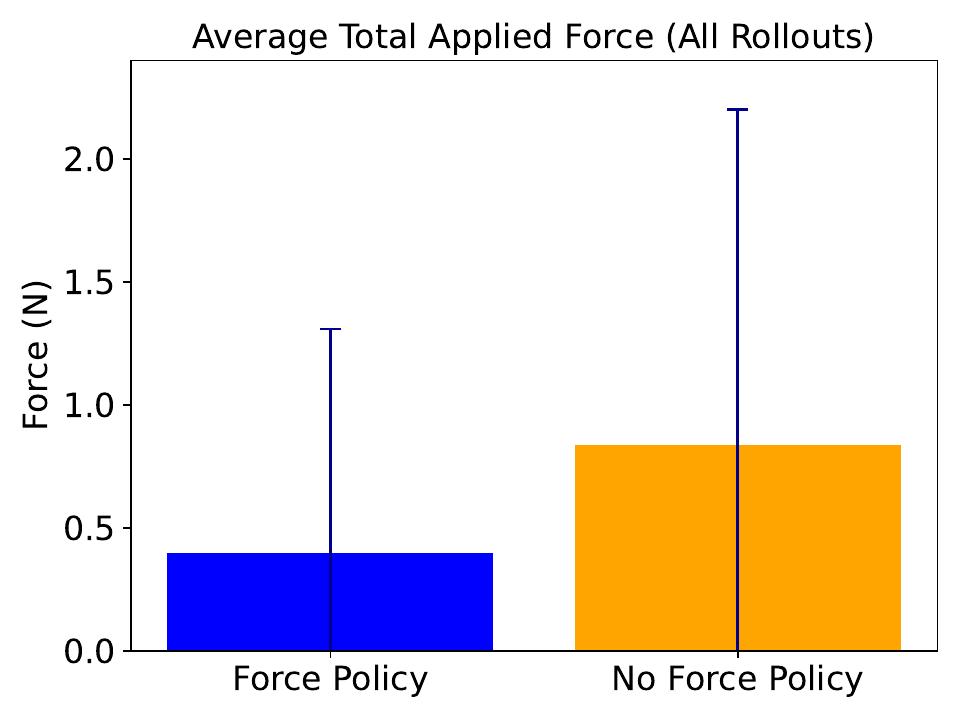}
        \caption{(a)}
        \label{fig:generalization_bar_plot}
    \end{subfigure}
    \hfill
    \begin{subfigure}[b]{0.32\textwidth}
        \centering
        \includegraphics[width=\textwidth]{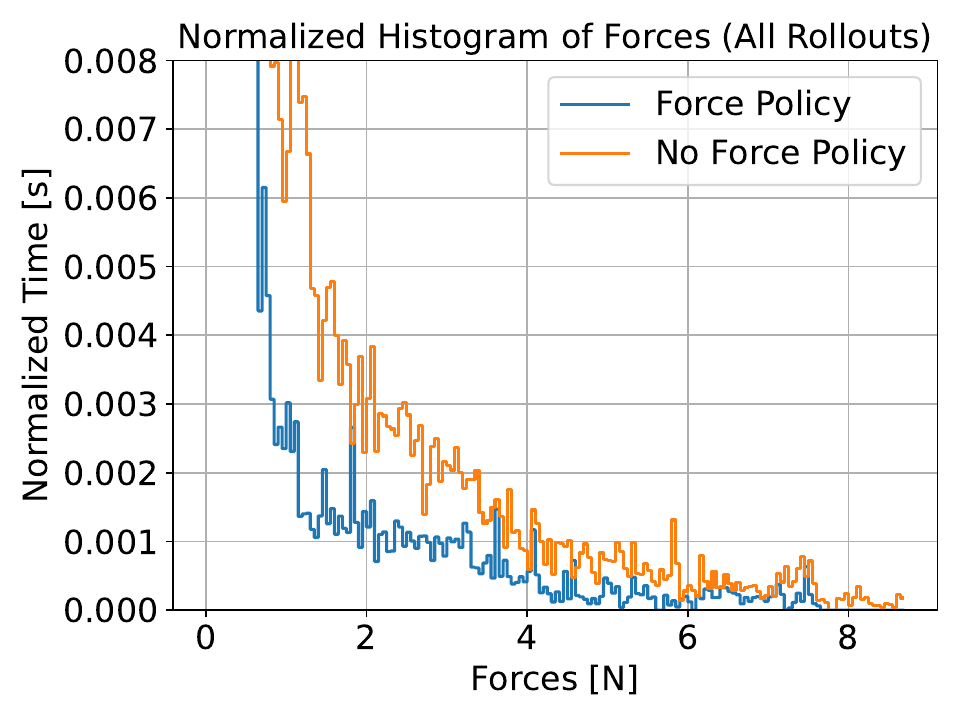}
      \caption{(b) }
      \label{fig:generalization_historgam_forces_plot}
    \end{subfigure}
    \hfill
    \centering
    \begin{subfigure}[b]{0.32\textwidth}
        \centering
        \includegraphics[width=\textwidth]{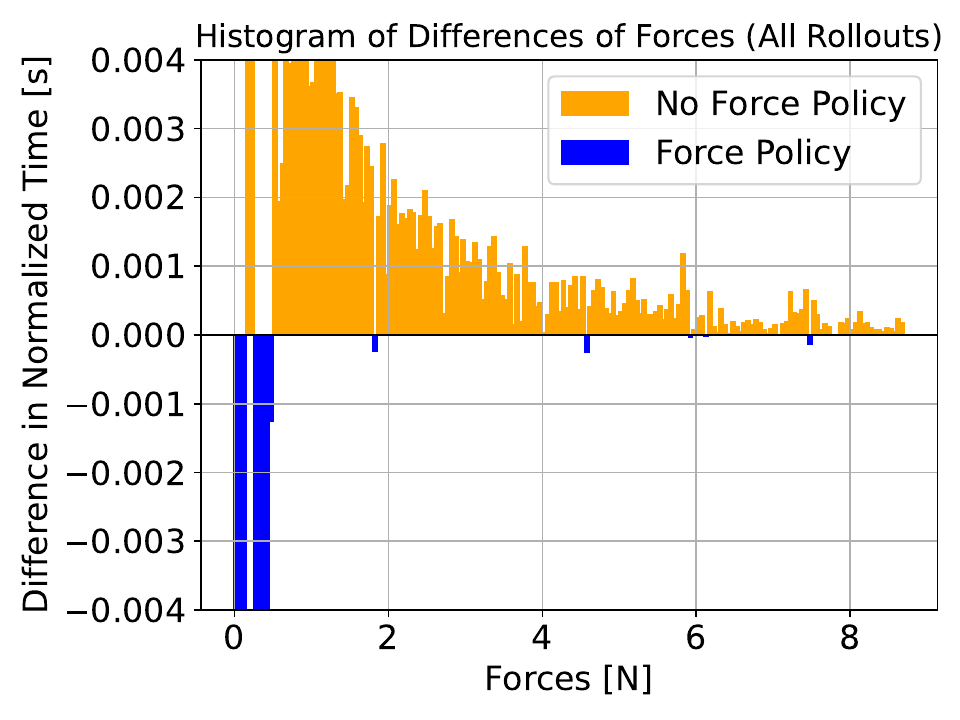}
      \caption{(c) } \label{fig:generalization_histogram_diff}
    \end{subfigure}
    \caption{(a) Average total applied force and standard deviation for both policy roll outs on the unseen tissue sample. (b) Normalized histogram of forces showing the total time a specific force is applied during both policy roll outs on the unseen tissue sample. (c) Histogram of differences of applied force during both policy roll outs on the unseen tissue sample.}
\end{figure*}

We also investigated the applied forces by the two polices on the unseen tissue sample. As shown in Fig.~\ref{fig:generalization_bar_plot}, the mean and standard deviation of the applied forces by the force policy are lower than those applied by the no force policy. In particular, our results show that the no force policy applies 110\% more force on average than the force policy. This result is also statistically significant. The statistical analysis of the above result using a t-test revealed a t-statistic of $-40.5$ and a p-value $<$ 0.01. This suggests that the force policy is also more gentle with the unseen tissue than the no force policy.

Fig.~\ref{fig:generalization_historgam_forces_plot} shows the distribution of the duration for which different forces were applied by each policy on the unseen tissue. The no force policy applied almost all the force values for longer durations, compared with the force policy. This also suggests that the force policy is more gentle with the tissue compared with no force policy. 

This is also illustrated in Fig.~\ref{fig:generalization_histogram_diff}, which shows the difference of the time a given force is applied between the two policies. In particular, our results show that the vast majority of forces applied more often by the force policy are less than 0.5N. In contrast, the no force policy applies larger forces more often. This provides additional evidence that the force policy is better at handling the unseen tissue sample throughout task execution compared with the no force policy.

\section{DISCUSSION}
       \label{sec:discussion}
       The above results confirm all our hypotheses in Section~\ref{sec:eval_metrics}. They show several benefits to include the tool-tissue interaction forces in the design of autonomous systems in RAS. These benefits include achieving higher success rate in autonomously performing the task and being more gentle with the tissue, compared with the force agnostic system. The results also show that those benefits are maintained when the force-aware autonomous system is applied to previously unseen tissue.

There are several potential reasons that justify the above benefits of using force data. For example, the force data indicates when the surgical instrument touches the tissue, providing the needed cue for the force-aware system to open the instruments' jaws to grasp the tissue correctly. In addition, the gentle manipulation of tissue can make it hard to detect any resulting tissue deformation using vision data only. Therefore, the lack of using force data can lead to undesirable situations.~On the one hand, the force-agnostic system may fail to grasp the tissue (and grasps the air instead) especially with changing the relative height of the tissue with respect to the surgical camera. On the other hand, the force-agnostic system may exert excessive force on the tissue when actually grasping it. This excessive force can make the tissue deformation clear in the vision data, giving the needed visual cue to the force-agnostic system to open the jaws and grasp the tissue. 

There are several ways to integrate our proposed method with clinical systems to realize the above benefits. This integration can be directly implemented with the newly released RAS systems with force feedback capabilities such as the da~Vinci~5 (Intuitive Surgical Inc.) and Hugo (Medtronic Inc.) systems. For clinical systems that do not have this capability, our proposed method can be integrated with methods from the literature that estimate the tool-tissue interaction forces~\cite{black20206, 10183676, chua2020toward}. In particular, the work in~\cite{chua2020toward} can be a good candidate since its ground truth data were generated using the same force/torque sensor used in our study. 

There are several directions to improve the current work. While the proposed force-aware system is better than the force-agnostic one, it still needs to meet the human demonstrator's performance. This includes matching the human's success rate (100\%) and applied force (average of 0.14 $\pm$ 0.3 N). Moreover, there are several other surgical tasks that can be used to validate the results of this work including suturing and dissection. 
        
\section{CONCLUSION AND FUTURE WORK}
        \label{sec:conclusion}
        In this work, we demonstrated the benefits of using tool-tissue interaction forces in designing autonomous systems in RAS. Our results show that force-aware autonomous systems have higher success rate and are more gentle with tissue compared with force-agnostic ones in the autonomous execution of a RAS manipulation task. Furthermore, our results show that  force-aware autonomous systems can better generalize to previously unseen tissue compared with force-agnostic ones. These results show that designing force-aware autonomous systems can lead to autonomous task execution that better adheres to surgical guidelines in handling the different tissue types in any surgical procedure.   

This work opens up several areas for future work. For example, it would be interesting to validate the findings of this work on other surgical tasks that involve physical interaction with the tissue. Furthermore, studying the effect of using different force estimation methods on the performance of the force-aware autonomous system can provide useful design insights. Another interesting direction is to use demonstrations from more than one demonstrator and explore how to account for style changes between them to maintain and improve the performance of the force-aware autonomous systems.  
        
\section*{ACKNOWLEDGMENTS}
        \label{sec:acknowledgement}
        We would like to thank Mary Kate Gale, Paula Stocco, Ghadi Nehme and Brian Vuong for their assistance in the early stages of this work.

\bibliographystyle{IEEEtran}
\bibliography{bibliography_force_project}

\end{document}